\documentclass[conference]{IEEEtran}
\IEEEoverridecommandlockouts
\usepackage{cite}
\usepackage{amsmath,amssymb,amsfonts}
\usepackage{graphicx}
\usepackage{textcomp}
\usepackage{xcolor}

\usepackage{booktabs}  
\usepackage{algpseudocode}
\usepackage{algorithm}

\def\BibTeX{{\rm B\kern-.05em{\sc i\kern-.025em b}\kern-.08em
    T\kern-.1667em\lower.7ex\hbox{E}\kern-.125emX}}
\begin{document}

\title{T\textsuperscript{2} of Thoughts: Temperature Tree Elicits Reasoning in Large Language Models\\
\thanks{\(^{*}\) Equal Contribution. \quad \(^{\dag}\) Corresponding Author.}
}

\author{\IEEEauthorblockN{\textsuperscript{*}Chengkun Cai}
\IEEEauthorblockA{\textit{School of Informatics} \\
\textit{Universiy of Edinburgh}\\
un1a4ting@gmail.com}
\and
\IEEEauthorblockN{\textsuperscript{*}Xu Zhao}
\IEEEauthorblockA{\textit{School of Informatics} \\
\textit{Universiy of Edinburgh}\\
shiningdn02@gmail.com}
\and
\IEEEauthorblockN{Haoliang Liu}
\IEEEauthorblockA{\textit{Department of EEE} \\
\textit{Universiy of Manchester}\\
mortalneo@outlook.com}
\and
\IEEEauthorblockN{Yucheng Du}
\IEEEauthorblockA{\textit{School of Informatics} \\
\textit{Universiy of Edinburgh}\\
Y.Du-53@sms.ed.ac.uk
}
\and
\IEEEauthorblockN{\textsuperscript{\dag}Lei li}
\IEEEauthorblockA{\textit{Computer Science Department} \\
\textit{University of Copenhagen}\\
lilei@di.ku.dk}
}

\maketitle

\begin{abstract}
Large Language Models (LLMs) have emerged as powerful tools in artificial intelligence, especially in complex decision-making scenarios, but their static problem-solving strategies often limit their adaptability to dynamic environments. We explore the enhancement of reasoning capabilities in LLMs through Temperature Tree (T\textsuperscript{2}) prompting via  a heuristic algorithm, termed as T\textsuperscript{2} of Thoughts (T\textsuperscript{2}oT). The primary focus is on enhancing decision-making processes by dynamically adjusting search parameters, especially temperature, to improve accuracy without increasing computational demands. We empirically validate that our hybrid T\textsuperscript{2}oT approach yields enhancements in, single-solution accuracy, multi-solution generation and text generation quality. Our findings suggest that while dynamic search depth adjustments based on temperature adaption can yield mixed results, a fixed search depth, when coupled with T\textsuperscript{2}oT’s adaptive capabilities, provides a more reliable and versatile problem-solving strategy. This work highlights the potential for future explorations in optimizing algorithmic interactions with foundational language models, particularly illustrated by our development for the Game of 24 and Creative Writing tasks. 
\end{abstract}


\section{Introduction}
Large Language models (LLMs) are increasingly employed across a broad spectrum of Natural Language Processing (NLP) tasks, including machine translation \cite{zhu2023multilingual}, summarization \cite{pang2023long}, and question answering \cite{zou2023representation}. In Recent years, a number of language models such as GPT \cite{radford2018improving,radford2019language,brown2020language, achiam2023gpt}, LLaMa \cite{touvron2023llamaA,touvron2023llamaB} are developed in a rapid speed. However, they are typically constrained to sequential, token-by-token processing during reasoning. These methods, while efficient, may be insufficient in scenarios that require forward-thinking or strategic planning, where early decisions significantly impact results. 

\begin{figure*}[h]
   \centering
   \includegraphics[width=\linewidth]{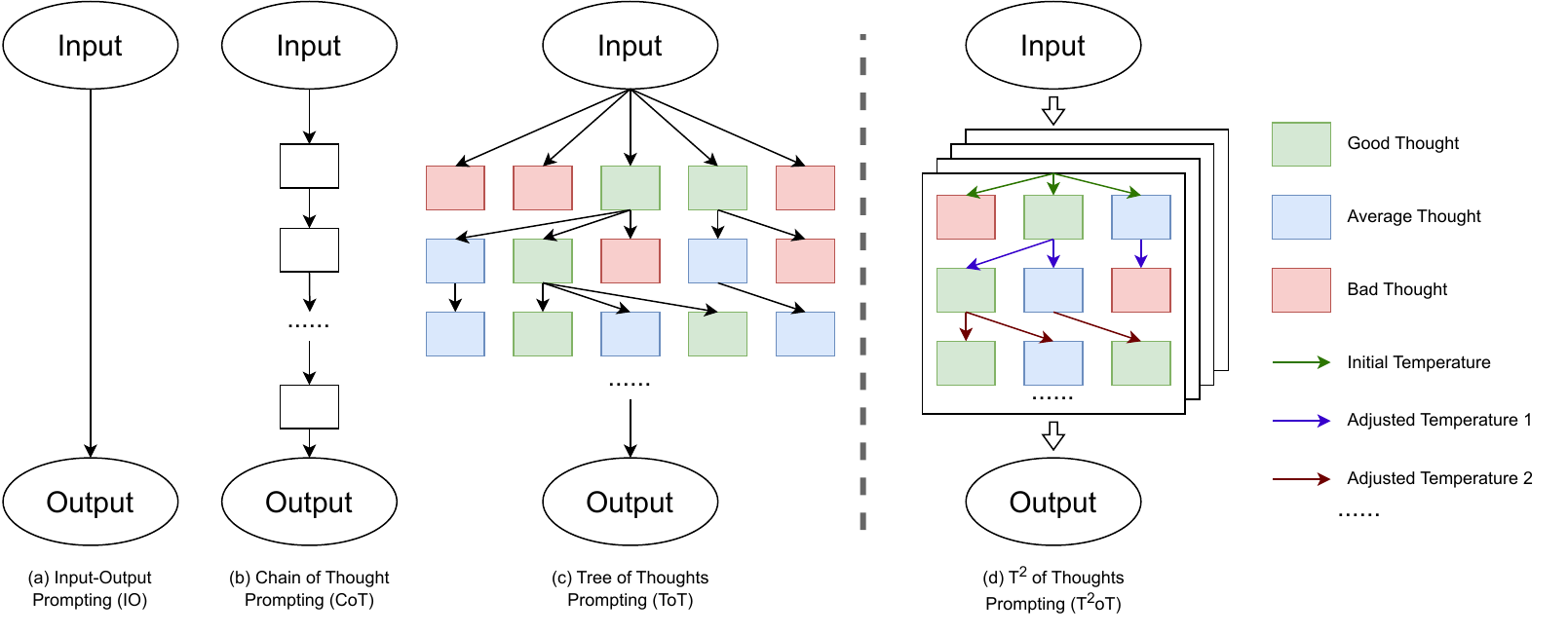}
   \caption{Our T\textsuperscript{2}oT compared with IO, CoT and ToT. Rectangles of different colors represent thoughts with different evaluations. Arrows of different colors represent different temperatures in the reasoning process.}
   \label{fig:compare}
\end{figure*}

Traditional approaches to enhancing these models, such as, Input-output (IO), Chain of Thought (CoT) \cite{wei2022chain}, have made strides by enabling models to follow a logical sequence of reasoning steps. However, this method often lacks the flexibility to explore multiple reasoning pathways concurrently or revisit prior decisions to optimize outcomes. To address these limitations, a new processing framework for language models, termed Tree of Thoughts (ToT) \cite{yao2024tree}, has been developed. Building on the well-regarded CoT, ToT enables language models to explore multiple reasoning pathways and evaluate various options to decide the next steps. This methodology not only anticipates future actions but also allows revisiting prior decisions to optimize overall outcomes, significantly enhancing the problem-solving capabilities of language models in complex tasks~\cite{li2024cpseg,shi2024chops}, such as Game of 24, Creative Writing and Mini Crosswords.

However, in these prompting methods, a significant parameter, temperature, is often set as fixed \cite{ouyang2023llm, chiang2023can}, which may introduce a number of limitations. To control the model generation, temperature sampling \cite{ackley1985learning} is used to control the decoding process and influence the model performance by adjusting the probability distribution of the next token to be generated. To achieve a balance between generation quality and diversity, temperature sampling should be used properly \cite{nasir2023llmatic}. In LLMs, the temperature primarily controls the randomness of text generation. A lower temperature value makes the model more likely to choose the conservative paths, leading to more deterministic and consistent outputs. Conversely, a higher temperature value increases the likelihood of selecting aggressive paths, resulting in more diverse and unpredictable text. As a result, fixed temperature restricts the flexibility to modulate the randomness and diversity of the responses based on different tasks. It has been demonstrated that using a fixed temperature is not the optimal choice in many cases, regardless of the type of language tasks being executed by the model \cite{zhang2024edt}.

As a result, a reasonable dynamic temperature adjustment strategy could be developed for problem-solving in LLMs. Some strategies of dynamic temperature sampling have been developed to improve the performance of LLMs~\cite{zhang2024edt,chang2023kl,zhu2024hot}. However, the strategies proposed by these works have some limitations, lack comprehensive analysis and are not connected with prompting methods in LLMs. Our research proposes T\textsuperscript{2} of Thoughts (T\textsuperscript{2}oT), a novel prompting method that addresses existing limitations by merging the structured exploration capabilities of Tree of Thoughts (ToT) with the dynamic adaptability of temperature adjustment. This integration enables more deliberate and refined reasoning processes.

Figure \ref{fig:compare} illustrates the architecture of our T\textsuperscript{2}oT compared with other prompting methods. Inspired from Particle swarm optimization (PSO) \cite{488968}, T\textsuperscript{2}oT can derive multiple trees. For each tree, it adjusts the temperature of the next reasoning step based on its own and other trees' evaluations of the previous reasoning steps.This method allows combining the own and the group's cognition of reasoning. 

The temperature adjustment based on the best evaluations from multiple trees allows T\textsuperscript{2}oT to correct insufficient reasoning directions, thereby significantly improving the overall quality of the reasoning process. The randomness introduced by the dynamic adjustment mechanism enhances the robustness of the method. Furthermore, T\textsuperscript{2}oT can better adapt to dynamically changing environments or problems, as the information from all trees is updated in real time.

Our contribution can be included for two points:

\begin{itemize}
    \item Dynamic Temperature Adjustment: We introduce T\textsuperscript{2}oT to dynamically adjust the temperature parameter during inference. This adjustment improves both the accuracy and diversity of solutions generated by GPT-4 \cite{achiam2023gpt}.
    \item Empirical Validation: We empirically validate our T\textsuperscript{2}oT by comparing it against ToT on GPT-4. Our results show significant enhancements in both single-solution accuracy and multi-solution generation in Game of 24 and improvement of coherency scores in Creative Writing.
\end{itemize}

\section{Related Work}
\paragraph{Prompting and Reasoning for LLMs.} 
Machine learning for prompting involves using various techniques to improve the ability of language modes to interpret and generate text based on given prompts. In recent years, research in this area has primarily focused on improving the efficiency and effectiveness of language models in understanding and responding to prompts. Various prompt engineering strategies are discussed and highlighted their importance in fine-tuning model behavior\cite{liu2023pre}.  A prompt-based fine-tuning and automatic prompting generation method are introduced to conduct few-shot fine-tuning of language models \cite{gao2021making}. In the research and application of LLMs, different prompting methods are developed and used to optimize the model's reasoning and generation capabilities. Input-Output (IO) prompting is the most basic method, where a specific input is provided, and the LLM generates the corresponding output. CoT breaks down complex reasoning tasks into a series of intermediate steps, allowing the model to reason step-by-step, building each step upon the previous one. ToT structures the reasoning process as a tree, where each node represents a potential solution path, allowing for making decisions of the next step by multiple paths and self-evaluation. Graph of Thoughts (GoT) \cite{besta2024graph} models the reasoning of a LLM as an arbitrary graph, where units of information  are considered as vertices, and dependencies between these vertices are considered as edges. This method allows for the integration of various LLMs thoughts to create synergistic results, to extract key insights from comprehensive thought networks, or to enhance thoughts through iterative feedback. Hypergraph of Thought (HoT) \cite{yao2023thinking} is applied in multimodal reasoning. This method uses triple as the primary thought to model higher-order relationships, thus generate a hyperedge of thought by multi-hop walking to achieve multi-hop inference.

\paragraph{Heuristic Optimization.} 

Artificial intelligence is closely related to heuristic algorithms, which often solve optimization and search problems by simulating the behavior of nature and living organisms. Evolutionary algorithms \cite{storn1997differential,vesterstrom2004comparative,brest2006self,zhang2009jade} are a type of optimization algorithm that imitates biological evolution mechanisms, whose core idea is to continuously evolve better solutions through an iterative process among the population of candidate solutions. The idea of evolutionary algorithms is used to optimize discrete prompt to have a better performance of LLMs \cite{guo2023connecting}. Particle swarm optimization (PSO)\cite{488968} is a method for optimization of continuous nonlinear functions. PSO operates on a population of potential solutions, called particles, which move through the solution space to find the optimal solution. Each particle represents a candidate solution and adjusts its position based on its own experience and the experience of neighboring particles. The particles track their own best positions (personal best) and the overall best position (global best) found by any particle in the swarm. The movement of the particles is influenced by their own past performance and the performance of their neighbors, guiding them towards better solutions over time. PSO has demonstrated strong performance in various application. For instance, in neural network training, PSO has been shown to be as effective as the traditional error back-propagation method in optimizing network weights \cite{fisher1936use}.

\section{Methodology}
\paragraph{Problem Formulation.}

Considering an LLM tasked with generating solutions in a dynamic environment, the reasoning approaches generate multiple ideas or hypotheses from the initial input, forming a treelike structure. Each node represents a thought process, and the branches represent the exploration of subsequent thoughts. 
However, in ToT, the temperature parameter, which controls the randomness of text generation, remains fixed. This static nature restricts the flexibility needed to adapt to varying contexts or needs.

To address this, we propose T\textsuperscript{2}oT for problem solving in Figure~\ref{fig:compare}. The primary objective is to dynamically adjust the temperature parameter, thereby modulating the randomness and diversity of the responses based on the reasoning state. This dynamic adjustment is governed by PSO, which leverages personal best and global best values to fine-tune the temperature.  Practically, T\textsuperscript{2}oT can derive multiple trees. For each tree, it adjusts the temperature at each step of the reasoning process based on its own best evaluation value (personal best) and the best evaluation value among all the trees (global best).  The problem can be formally defined as follows:

\begin{itemize}
    \item \textbf{Objective:} Enhance the problem-solving capabilities of LLMs by dynamically adjusting the temperature parameter using T\textsuperscript{2}oT.
    \item \textbf{Constraints:} Maintain computational efficiency equivalent to ToT when limiting the number of number of trees to one.
\end{itemize}

The temperature adjustment can be expressed as:

\begin{equation}
\begin{aligned}
T_i[n] & = w_0 \cdot T_i[n-1] 
\\
& + \alpha_1 \cdot (pb_i[n-1] - x_i[n])
\\
& + \alpha_2 \cdot (gb[n-1] - x_i[n])
\end{aligned}
\label{Equ:tree}
\end{equation}

In the Equation~\ref{Equ:tree}, $n$ represents the n-th reasoning step, $T$ denotes the temperature, $i$ is the index of the current tree, $pb$ stands for the personal best value, $gb$ represents the global best value, $x$ indicates the evaluation value, $w_0$ is the inertial weight, and $\alpha_1$ and $\alpha_2$ are the acceleration coefficients for personal and global bests, respectively.

The inertial weight governs the degree to which the current temperature is influenced by the previous temperature, thereby affecting the exploration-exploitation balance in the reasoning process. The acceleration coefficient for the personal best dictates the extent to which the temperature is adjusted towards the local optimal solution identified by the individual tree. This parameter enhances the algorithm's ability to utilize its accumulated knowledge. Conversely, the acceleration coefficient for the global best determines the extent to which the temperature is adjusted towards the global optimal solution identified across all trees, thereby emphasizing the collective learning from the group's experience.

The adjusted temperature $T_{\text{n}}$ is subsequently constrained within a predefined range to ensure stability and to prevent excessive randomness or determinism, which could undermine the effectiveness of the reasoning process.


\paragraph{T\textsuperscript{2} of Thoughts.}

 \begin{algorithm}[H]
    \caption{T\textsuperscript{2}oT Temperature Adjustment in a Tree}
    \begin{algorithmic}[1] 
        \State \textbf{Input:} Current reasoning step $n$, temperature in previous reasoning step $T[n]$, $x[n]$, $pb[n-1]$, $gb[n-1]$,  maximum temperature limit $T_{max}$, minimum temperature limit $T_{min}$, maximum reasoning step $N$.
        \State \textbf{Output:} Adjusted temperature in n-th reasoning step in a particular tree.
        \For{n in range(0, N)}   
            \State $T[n] \leftarrow w_0 \cdot T[n-1] + \alpha_1 \cdot (pb_i[n-1] - x[n]) + \alpha_2 \cdot (gb[n-1] - x[n])$
            \If{$T_n > T_{max}$}
                \State $T_n \leftarrow T_{max}$
            \ElsIf{$T_n < T_{min}$}
                \State $T_n \leftarrow T_{min}$
            \EndIf
        \EndFor
    \end{algorithmic}
    \label{Alg:T}
\end{algorithm}

\begin{figure*}[ht]
   \centering
   \includegraphics[width=\linewidth]{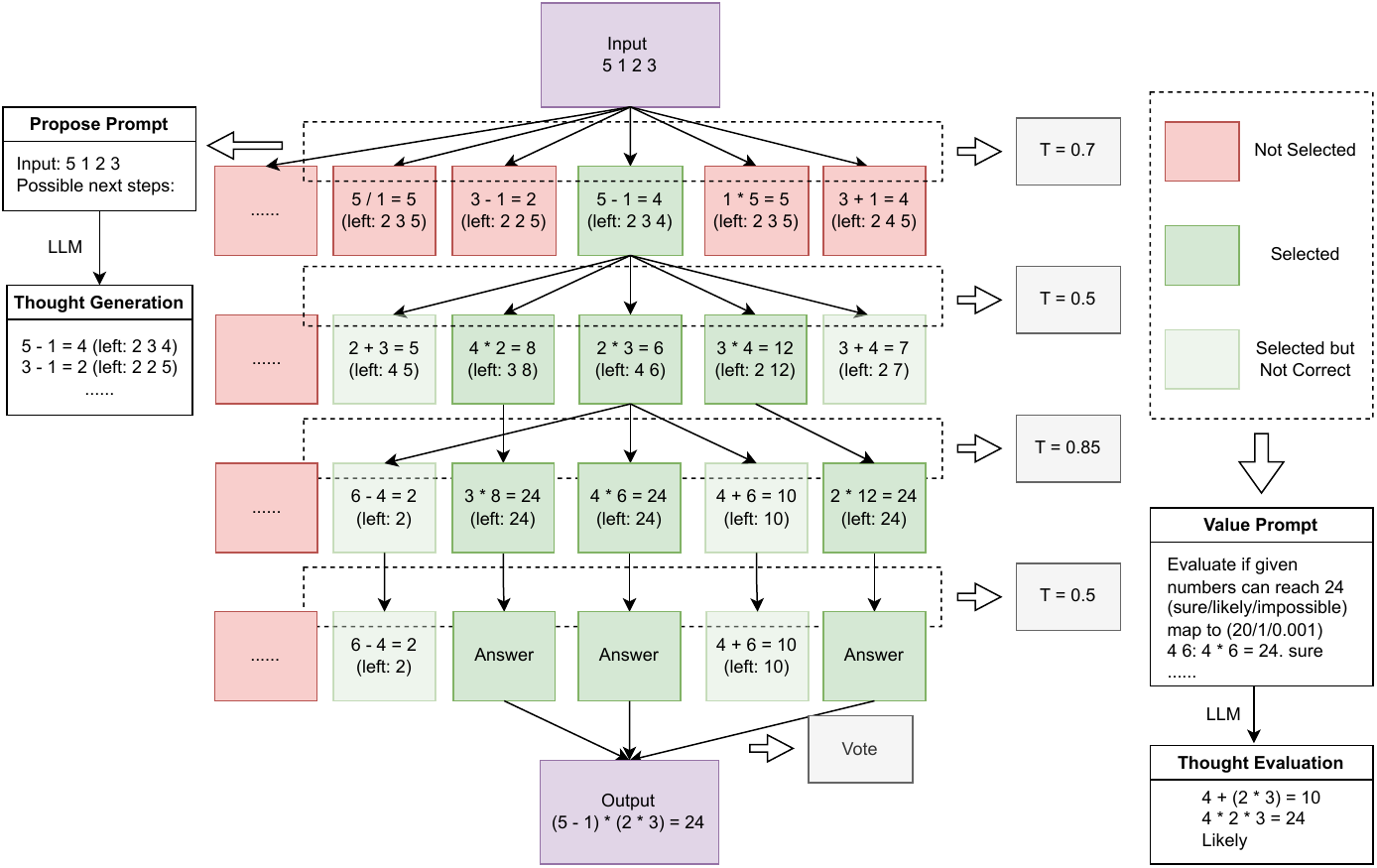}
   \caption{T\textsuperscript{2}oT in a Game of 24. The LM is prompted for thought generation and evaluation. The temperature changes in each step of reasoning. }
   \label{fig:Gameof24}
\end{figure*}

The T\textsuperscript{2}oT framework efficiently solves the temperature optimization problem by dynamically adjusting the temperature scale at each iteration, as described in Algorithm~\ref{Alg:T}. This iterative process ensures that the pipeline converges to the correct temperature \(T^*\) while avoiding significant errors. Through careful tuning of hyperparameters \(\alpha_1\), \(\alpha_2\), and \(w_0\), the framework balances local and global factors, enabling accurate and efficient convergence to the optimal solution with minimal error.

\section{Experiment}

\subsection{Game of 24}
\subsubsection{Experimental Setup}
\paragraph{Task.}
To evaluate our method, we employed the same task as used in ToT: Game of 24 in GPT-4. This game involves using arithmetic operations to manipulate four numbers to arrive at the number 24, a common benchmark in studies of computational problem solving and algorithmic efficiency. This specific task selection helps in directly comparing the performance and efficacy of T\textsuperscript{2}oT against established results in ToT, providing a clear metric for improvement or competitiveness in problem-solving scenarios. To evaluate the enhancements of T\textsuperscript{2}oT, we conducted two distinct types of experiments. The first experiment is designed to evaluate the performance of our method across 50 games on each method, comparing the accuracy of T\textsuperscript{2}oT against ToT. The second experiment focused specifically on T\textsuperscript{2}oT's capability to explore multiple solutions. We selected a specific instance of a game "7, 5, 2, 6" which have multiple solutions and run both methods on this instance 10 times each and track the frequency of distinct solutions produced by each run.

\paragraph{Dataset.}
In order to set up a reliable experiment plan for T\textsuperscript{2}oT, to test its performance when applied into a large language model, a newly generated dataset is necessary. To align with our experiments, the dataset for this task was algorithmically generated. Each instance of the Game of 24 consists of four numbers.


\paragraph{Baseline and Framework Setup.}
For our experiments, we use ToT as the baseline and the same setup in our T\textsuperscript{2}oT. T\textsuperscript{2}oT breaks down the thought process into three steps. We set the depth of search to 3.  Utilizing a breadth-first search (BFS) methodology within the framework, we consistently select and advance the top five most promising solutions at every stage of the decision-making process. As shown in Figure \ref{fig:Gameof24}, this selection is evaluated by the language model prompt, which classifies each potential solution into categories: "sure", "maybe" or "impossible" based on the likelihood of reaching the total of 24. The primary objective remains to progress viable partial solutions that can be substantiated with a limited number of predictive steps, eliminate those solutions that are evidently unfeasible through practical judgments such as being "too large" or "too small," and continue to explore those deemed as "maybe". We perform value sampling three times for each conceptual step in the thought process.

\begin{figure*}[ht]
   \centering
   \includegraphics[width=\linewidth]{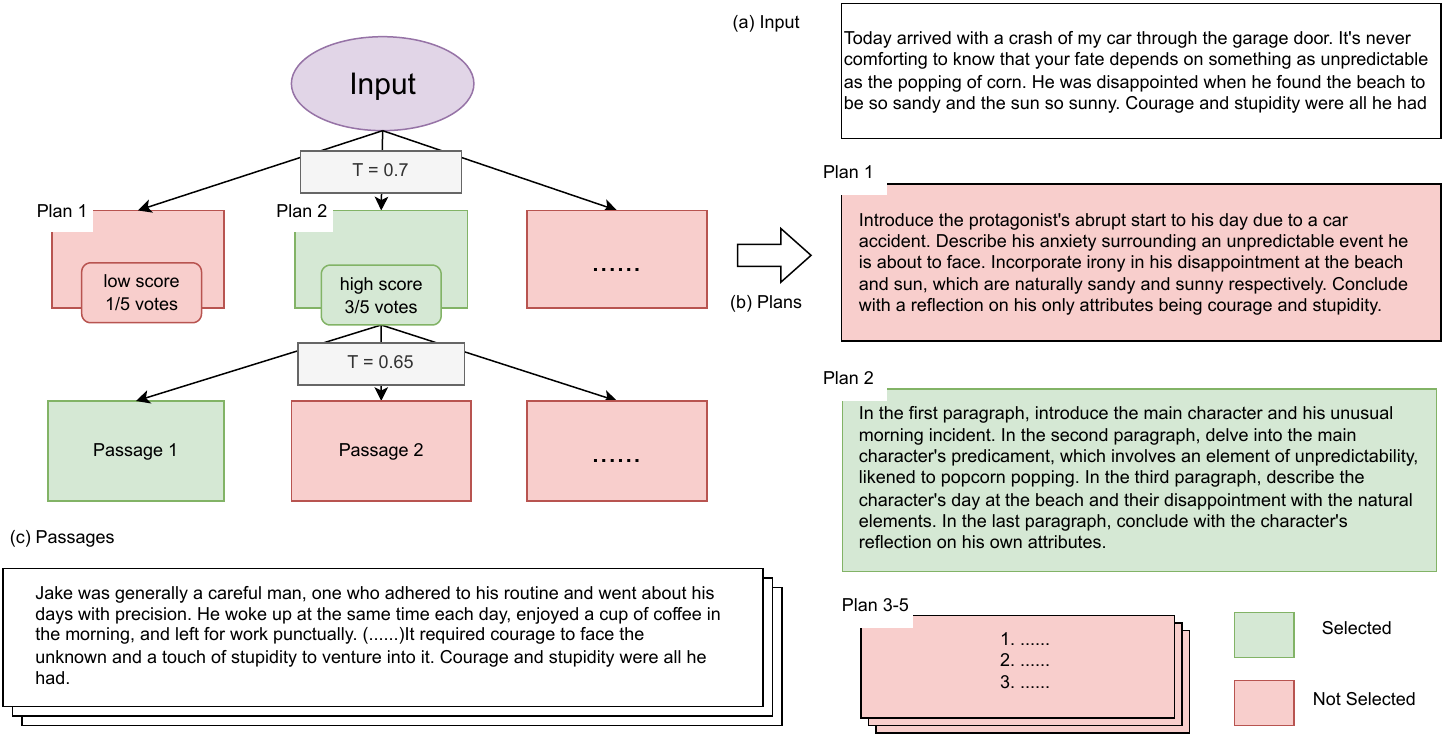}
   \caption{T\textsuperscript{2}oT in Creative Writing. After generating five plans by prompting, dynamically adjust the temperature for producing passages based on the highest score.}
   \label{fig:CreaitiveWriting}
\end{figure*}

\paragraph{Parameter Setup.}
In this experiment, the inertial weight was set to 1, as each temperature update is based on the previous temperature, making it reasonable to set the inertial weight to 1. The acceleration coefficients for personal best and global best were both set to 0.1 for two reasons. First, this normalization is based on the magnitude of the temperature values and the evaluation scores for each node. Second, we assume that the influence of each individual step of reasoning and the overall influence of the group are equal. The number of trees in T\textsuperscript{2}oT is another hyperparameter that can be adjusted. Increasing this parameter is equivalent to performing multiple ToT reasoning for the same input in terms of computational efficiency. In the experiments, we set the number of trees to one to control that the computational efficiency of T\textsuperscript{2}oT is the same as ToT.

\subsubsection{Results}

\begin{table}[h]
  \centering
  \caption{Game of 24 Success Rates}
  \label{tab:algorithm_comparison_24}
  \begin{tabular}{lc}
    \toprule
    Method & Success \\
    \midrule
    IO & 7.3\% \\
    CoT & 4.0\% \\
    ToT & 74\% \\
    T\textsuperscript{2}oT (ours) & 80\% \\
    \bottomrule
  \end{tabular}
\end{table}

\begin{table}[htbp]
  \centering
  \caption{Game of 24 Multiple Solutions Results}
  \label{tab:algorithm_comparison_solutions}
  \resizebox{\columnwidth}{!}{%
  \begin{tabular}{lcc}
    \toprule
    \textbf{Method} & \textbf{Solution Types} & \textbf{Solution Frequency} \\
    \midrule
    ToT & 2 & (0.9, 0.1) \\
    T\textsuperscript{2}oT (ours) & 3 & (0.5, 0.3, 0.2) \\
    \bottomrule
  \end{tabular}%
  }
\end{table}

As shown in Table \ref{tab:algorithm_comparison_24}, IO and CoT prompt methods perform badly on this task. The success rate for T\textsuperscript{2}oT is 80\%, which is higher than ToT at 72\%. This indicates an improvement of 8\% increase in success.  T\textsuperscript{2}oT appears to provide a more effective search strategy than ToT. The increased accuracy suggests that T\textsuperscript{2}oT may be better at navigating the problem space or more efficient at finding solutions that meet the goal. This could be due to better exploration of the search space or more effective pruning of less promising paths.

The result of the second experiment is shown in Table \ref{tab:algorithm_comparison_solutions}. ToT is able to generate two distinct solution types for the given game, with the primary solution type occurring with a high frequency of 0.9 and a secondary, less frequent type at 0.1. In contrast, T\textsuperscript{2}oT demonstrates a superior ability to diversify its approach, producing three different types of solutions with a more balanced distribution of frequencies at 0.5, 0.3, and 0.2, respectively. This indicates that T\textsuperscript{2}oT not only increases the number of potential solutions explored but also promotes a more equitable exploration across these solutions, preventing over-concentration on a dominant solution path.

\subsection{Creative Writing}
\subsubsection{Experimental Setup}

\begin{figure*}[htbp]
   \centering
   \includegraphics[width=1.0\linewidth]{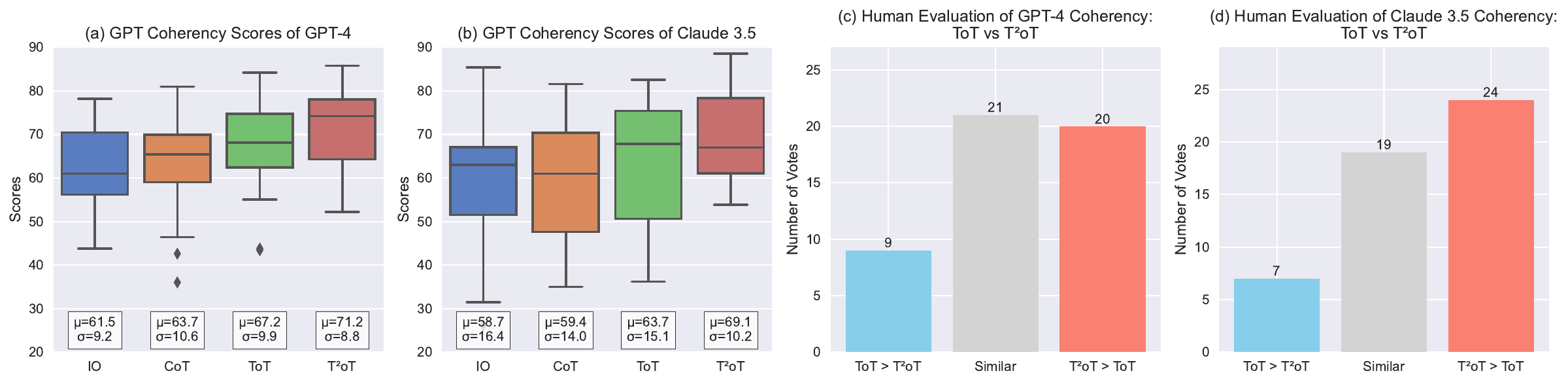}
   \caption{Creative Writing results.} 
   \label{fig: Creative Writing results}
\end{figure*}

\paragraph{Task.}
To evaluate our algorithm, we employed the same Creative Writing task as used in ToT \cite{yao2024tree} on GPT-4. In this task, 4 random sentences are given as input. The goal is to produce a coherent passage divided into 4 paragraphs, with each paragraph concluding with one of the given sentences. This open-ended task encourages creative thinking and complex planning. We use 50 sentence inputs, which are sampled randomly from randomwordgenerator.com, with no predefined correct passage for each set of constraints.

\paragraph{Baseline and Framework Setup.} For our experiments, we also utilize ToT as the baseline, employing a comparable setup in our T\textsuperscript{2}oT. As shown in Figure \ref{fig:CreaitiveWriting}, T\textsuperscript{2}oT breaks down the thought process into two steps (depth = 2): Firstly, the language model generates 5 different plans, then conducts 5 rounds of voting to determine the best plan. Secondly, the majority choice is used to write the output passage, following the same generate-and-vote procedure.
As we find that GPT can follow the input constraints most of the time, and since the original evaluation method can lead to inflated scores in the first step of plan assessment, and the 10-point scale results in many similar scores, making it difficult to distinguish among plans, we use a GPT zero-shot prompt to provide a 1-100 scalar score: "Analyze the following passage in detail. Consider the clarity, structure, argument coherence, and style of the writing. (......) Conclude with: 'Thus, the coherency score is {s}', where s is an integer from 0 to 100, aiming for a normal distribution of scores with an average of 50." This evaluation method aims for a normal distribution of scores with an average score of 50. 

\paragraph{Parameter Setup.}
In this experiment, the inertial weight is also set to 1, as each temperature update is based on the previous temperature, making it reasonable to set the inertial weight to 1. The initial temperature for all experiments is set to 0.7, as this is considered a balanced temperature. The acceleration coefficients for personal best and global best were both set to -0.005. we still set the number of trees to one to ensure that the computational efficiency of T\textsuperscript{2}oT are the same as ToT.

\subsubsection{Results}


Figure \ref{fig: Creative Writing results}(a) presents the average GPT-4 coherency scores across various experimental setups in GPT-4-0613 , illustrating that T\textsuperscript{2}oT, with an average score of 71.4, produces more coherent passages compared to IO (62.28), CoT (64.8) and ToT (67.5). While Figure \ref{fig: Creative Writing results}(b) illustrates that in Claude-3.5-haiku-20241022, our T\textsuperscript{2}oT has an average score of 68.3, which is also higher than IO (58.7), CoT (59.4) and ToT (63.7). It is worth noting that in the Creative Writing experiments conducted with Claude in Figure \ref{fig: Creative Writing results}(b), we still used GPT-4-0613 to evaluate the results to ensure consistency with the evaluation metric used in experiments conducted with GPT-4 in Figure \ref{fig: Creative Writing results}(a). The T\textsuperscript{2}oT appears to provide a more effective search strategy and offer a good dynamic balance: lowering the temperature to achieve higher textual consistency when plan consistency is low, and raising the temperature to enhance creativity when plan consistency is high. Figure \ref{fig: Creative Writing results}(c) confirms the trend by showing that in the experimets of GPT-4, humans prefer the results generated by ToT over T²oT in 9 out of 50 passage pairs, while preferring T²oT over ToT in 20 pairs, with the remaining 21 pairs judged as similarly coherent. Figure \ref{fig: Creative Writing results}(d) shows that humans preferred the results generated by ToT in 7 pairs, while preferring T²oT in 24 pairs, with the remaining 19 pairs judged as similarly coherent.

\begin{table}[h]
    \centering
    \caption{Evaluation with non-default temperature ToT}
    \label{tab:random_temperature}
    \resizebox{\columnwidth}{!}{%
    \begin{tabular}{lc}
        \hline
        \textbf{Method} & \textbf{Score (Mean ± Std)} \\
        \hline
        IO & 61.41 ± 9.48 \\
        CoT & 63.71 ± 11.00 \\
        ToT (random temperature) & 65.26 ± 10.36 \\
        ToT & 67.57 ± 10.31 \\
        T²oT & 71.59 ± 8.79 \\
        \hline
    \end{tabular}%
    }
\end{table}

Table \ref{tab:random_temperature} presents the results of using ToT with randomly sampled temperatures from the range (0,1) at each inference step, compared to default temperature ToT and our T²oT method. Since our approach dynamically adjusts the temperature during inference, this experiment demonstrates that random temperature adjustment does not lead to performance improvements. In contrast, the temperature adjustments made by our algorithm contribute to enhanced performance.

\section{Cost and efficiency}
Under the single-particle conditions used in this paper, the computation cost of using T\textsuperscript{2}oT is very similar to that of ToT. Table \ref{tab:Game_of_24_Cost} shows that in Game of 24, solving a problem with T\textsuperscript{2}oT and ToT both requires about 5.5k completion tokens, which is very similar to the tokens required by 100 CoT trials and selecting the best result. This shows that the number of tokens required to scoring the results in T\textsuperscript{2}oT is very small compared to the computation. 

\begin{table}[htbp]
  \centering
  \caption{Game of 24 Cost}
  \label{tab:Game_of_24_Cost} 
  \resizebox{\columnwidth}{!}{%
  \begin{tabular}{lcc}
    \toprule
    \textbf{Method} & \textbf{Generate/Prompt tokens} & \textbf{Cost per case} \\
    \midrule
    IO & 1.8k / 1.0k & 0.13 \\
    CoT & 6.7k / 2.2k & 0.47 \\
    ToT & 5.5k / 1.4k & 0.74 \\
    T\textsuperscript{2}oT(ours) & 5.5k / 1.6k & 0.80 \\
    \bottomrule
  \end{tabular}%
  }
\end{table}

\noindent In Creative Writing (Table \ref{tab:Creative_Writing_Cost} below), the number of completion tokens required by  T\textsuperscript{2}oT and ToT are also very similar. As a result, in our experiments, our single-particle T\textsuperscript{2}oT has a significant performance improvement than ToT without increasing the number of completion tokens and with less increase in cost.

\begin{table}[htbp]
  \centering
  \caption{Creative Writing Cost}
  \label{tab:Creative_Writing_Cost}
  \resizebox{\columnwidth}{!}{%
  \begin{tabular}{lcc}
    \toprule
    \textbf{Method} & \textbf{Generate/Prompt tokens} & \textbf{Cost per case} \\
    \midrule
    IO & 0.9k / 0.4k & 0.13 \\
    CoT & 0.9k / 0.4k & 0.47 \\
    ToT & 4k / 2.9k & 0.74 \\
    T\textsuperscript{2}oT(ours) & 4k / 3.8k & 1.01 \\
    \bottomrule
  \end{tabular}%
  }
\end{table}

\section{Discussion}
\paragraph{Multiple Trees in T\textsuperscript{2}oT.}
T\textsuperscript{2}oT supports setting the number of trees.  Setting multiple trees is equivalent to performing multiple ToT reasoning for the same input in terms of computational efficiency. In our experiments,  to control the computational efficiency of T\textsuperscript{2}oT is the same as ToT, this parameter is set to one. When there are multiple trees in T\textsuperscript{2}oT, the method is not only based on the historical experience of a single tree, but also learns from the successful experience of the group, thereby adapting to the solution of complex problems more flexibly and accurately. Since each tree is constantly updating and adjusting its own reasoning path, the reasoning  capabilities of the entire system will increase. The global best results in the group can help other trees correct deviated or insufficiently optimized inference directions, improving the quality of the solution as a whole.


\paragraph{Future Directions}
Future research could enhance the T\textsuperscript{2}oT framework by incorporating adaptive learning mechanisms for parameter optimization. Integrating neural networks into the reasoning process could enable the automatic adjustment of inertial weight and acceleration coefficients based on real-time performance feedback, transforming T\textsuperscript{2}oT into a more flexible and learnable algorithm, thereby reducing the need for manual parameter tuning. Additionally, exploring the application of T\textsuperscript{2}oT in other complex problem-solving domains, such as natural language processing tasks and multi-modal reasoning, such as HoT \cite{zhang2023multimodal}, could further validate its efficacy and versatility. Research could also investigate the scalability of T\textsuperscript{2}oT with larger datasets and more extensive computational resources, aiming to understand its potential in large-scale AI systems. Moreover, integrating reinforcement learning techniques to dynamically adjust the temperature parameter in response to changing task complexities could lead to more robust and adaptive AI solutions, advancing the state of the art in adaptive reasoning algorithms.

\section{Conclusion}
In this paper, we present T\textsuperscript{2}oT, a novel reasoning method that dynamically adjusts the temperature parameter during inference in large language models, significantly enhancing both the accuracy and diversity of solutions generated by GPT-4. Through comprehensive empirical validation, T\textsuperscript{2}oT demonstrates superior performance over the static temperature method ToT, achieving higher single-solution accuracy and improved multi-solution generation in tasks such as the Game of 24, as well as better coherency in Creative Writing. This integration of heuristic algorithms with artificial intelligence exemplifies a promising approach for developing more adaptive and efficient language model prompting techniques.

\section{Limitations}
While the proposed T\textsuperscript{2}oT algorithm demonstrates significant improvements in reasoning and solution diversity compared to traditional fixed-temperature methods, it still has several limitations. First, the manual tuning of hyperparameters such as inertial weight and acceleration coefficients can be suboptimal in different tasks. These fixed parameters may limit the adaptability of the algorithm in dynamic environments where real-time adjustments could lead to better performance.

Another limitation is the absence of a learnable mechanism for adjusting the temperature parameter. The current approach relies on a heuristic strategy based on particle swarm optimization (PSO), which, although effective, may not always provide the most efficient temperature adjustments across all problem domains. Future work could explore integrating reinforcement learning techniques or neural networks to automatically learn these hyperparameters based on feedback from the model's performance with multiple trees.

Furthermore, while T\textsuperscript{2}oT has demonstrated robust performance in structured tasks such as the Game of 24 and Creative Writing, its generalizability to more complex or less structured tasks, particularly those involving multimodal data, remains to be fully tested. The evaluation of its effectiveness in such domains would require further research.

\bibliographystyle{IEEEtran}
\bibliography{ref}


\end{document}